\documentclass[letterpaper, 10 pt, conference]{ieeeconf}  % Comment this line out if you need a4paper

\IEEEoverridecommandlockouts                              % This command is only needed if 
\usepackage{times}
\usepackage{cite}
\usepackage{multicol}
\usepackage[bookmarks=true]{hyperref}
\usepackage{amsmath,amssymb,amsfonts}
\usepackage{algpseudocode}
\usepackage{graphicx}
\usepackage{float}
\usepackage{subcaption}
\usepackage{textcomp}
\usepackage{xcolor}
\usepackage{balance}
\usepackage{lipsum}
\usepackage{cuted}
\let\labelindent\relax
\usepackage{enumitem}
\usepackage[english]{babel}
\usepackage{algorithm}
\usepackage{algpseudocode}
\newtheorem{theorem}{Theorem}

\newtheorem{prop}{Proposition}

\usepackage{booktabs}  % For nicer horizontal rules
\usepackage{array}     % For better column formatting
\usepackage{xcolor} % For coloring rows (optional)

\newtheorem{definition}{Definition}
\DeclareMathOperator*{\argmax}{arg\,max}

\title{\LARGE \bf
Quality Over Quantity: Curating Contact-Based Robot Datasets Improves Learning
}

\author{
    Hrishikesh Sathyanarayan$^1$, Victor Vantilborgh$^2$, and Ian Abraham$^{1}$ \\
    $^1$Department of Mechanical Engineering, Yale University, USA \\
    Emails: \{hrishi.sathyanarayan, ian.abraham\}@yale.edu \\
    $^2$Department of Electromechanical,
    Systems and Metal Engineering, Ghent University,
    Belgium \\
    Email: victor.vantilborgh@ugent.be \\
    \vspace{-2em}
}

\begin{document}

\maketitle
\thispagestyle{empty}
\pagestyle{empty}

%%%%%%%%%%%%%%%%%%%%%%%%%%%%%%%%%%%%%%%%%%%%%%%%%%%%%%%%%%%%%%%%%%%%%%%%%%%%%%%%
\begin{abstract}
    In this paper, we investigate the utility of datasets and whether more data or the ``right'' data is advantageous for robot learning.
    In particular, we are interested on quantifying the utility of contact-based data as contact holds significant information for robot learning. 
    Our approach derives a contact-aware objective function for learning object dynamics and shape from pose and contact data. 
    We show that the contact-aware Fisher-information metric can be used to rank and curate contact-data based on how \emph{informative} data is for learning. 
    In addition, we find that selecting a reduced dataset based on this ranking improves the learning task while also making learning a \emph{deterministic} process. 
    Interestingly, our results show that more data is not necessarily advantageous, and rather, less but \emph{informative} data can accelerate learning, especially depending on the contact interactions.
    % This paper demonstrates an automated method to curate and adapt contact-based datasets for robot learning. 
    % We show that more data is not necessarily advantageous, and rather, less but \emph{informative} data can accelerate learning, especially with contact-based interactions.
    % We show that more data is often not that advantageous, and that, in fact, often less data can improve learning, especially with contact-based interactions. 
    % We show that the metric is able to rank data according to their utility in improving the overall learning task. 
    % We show empirical evidence on the contact-nets dataset, as well as the Push-T diffusion model dataset. 
    Last, we show how our metric can be used to provide initial guidance on data curation for contact-based robot learning. 

\end{abstract}

%%%%%%%%%%%%%%%%%%%%%%%%%%%%%%%%%%%%%%%%%%%%%%%%%%%%%%%%%%%%%%%%%%%%%%%%%%%%%%%%
\section{INTRODUCTION}

    Data plays a crucial role across a wide variety of robot learning problems.
    % Data plays a crucial role in robot learning and training policies useful for a variety of manipulation problems.
    However, not all data is guaranteed to hold information that sufficiently informs the learning problem. 
    % the information richness in contact data is not always guaranteed.
    % Rather, contact interactions, while holding immense amounts of information useful for robot parameter learning problems, requires an active learning approach to seek information-rich contact seeking behaviors \cite{sathyanarayan2025behaviorsynthesiscontactawarefisher}. 
    The widely adopted convention is to collect more data, but without human-guided intuition~\cite{ARGALL2009469} or problem specific information~\cite{schaal_dmps}, it is exceedingly challenging, and often resource-intensive to generalize data collection in a formulaic method for robot learning. 
    This is especially true with systems that integrate tactile or contact-based data which has been shown to be highly informative \cite{sathyanarayan2025behaviorsynthesiscontactawarefisher}, but challenging to acquire~\cite{directposa}.
    % Therefore, in an attempt to make data more ubiquitous and streamlined in robotics, this paper studies what makes datasets more useful than others and establish methods that \emph{curates} data. 
    Therefore, in an attempt to make data more ubiquitous and streamlined in robotics, this paper challenges the 'more data is better' paradigm by providing a 'less is more' information metric to curate smaller, yet more powerful, robot learning contact-based datasets.
    
    Our focus is on studying contact-based data which has distinct identifiable features, e.g., was contact made or not and where.
    % This paper presents a method that regulates, ranks, and generates \emph{maximally informative} data across a variety of contact-based learning problems.
    % What makes contact interesting is that depending on how physical contact is made, one can isolate the physical parameters that can be learned from contact~\cite{nima}.
    Contact is particularly interesting because its multi-modal nature dictates the identifiability of different physical parameters \cite{nima}.
    As a guiding principle, we view parameter learning from an information theoretic approach in order to study what makes a good contact or not for learning.
    Inspired by data curation methods in \cite{agia2025cupidcuratingdatarobot}, we explore various measures of information, e.g., the Fisher information measure \cite{atanasov2013information,AF_Emery_1998,wilson_fishermax,sathyanarayan2025behaviorsynthesiscontactawarefisher}, that have been commonly used in robotics and adapt them to gain deeper intuition towards procedurally selecting the ``right'' data.

    % We show that a contact-aware formulation of the Fisher information allows us to rank data based on its utility in improving the numerical conditioning the learning problem.
    % Empirically, this approach to data curation yields a deterministic process for robot learning based on usefulness of data, rather than randomness. 
    
    % By optimizing over Fisher information, we generate information-rich data that is used to collapse on parameter uncertainty \cite{AF_Emery_1998,fedorov2010optimal,kiefer1959}.
    % We can therefore use the contact-aware Fisher information \cite{sathyanarayan2025behaviorsynthesiscontactawarefisher} to examine data utility in existing contact-based parameter learning problems.
    
    \begin{figure}[!t]
        \centering
        \includegraphics[width=\linewidth]{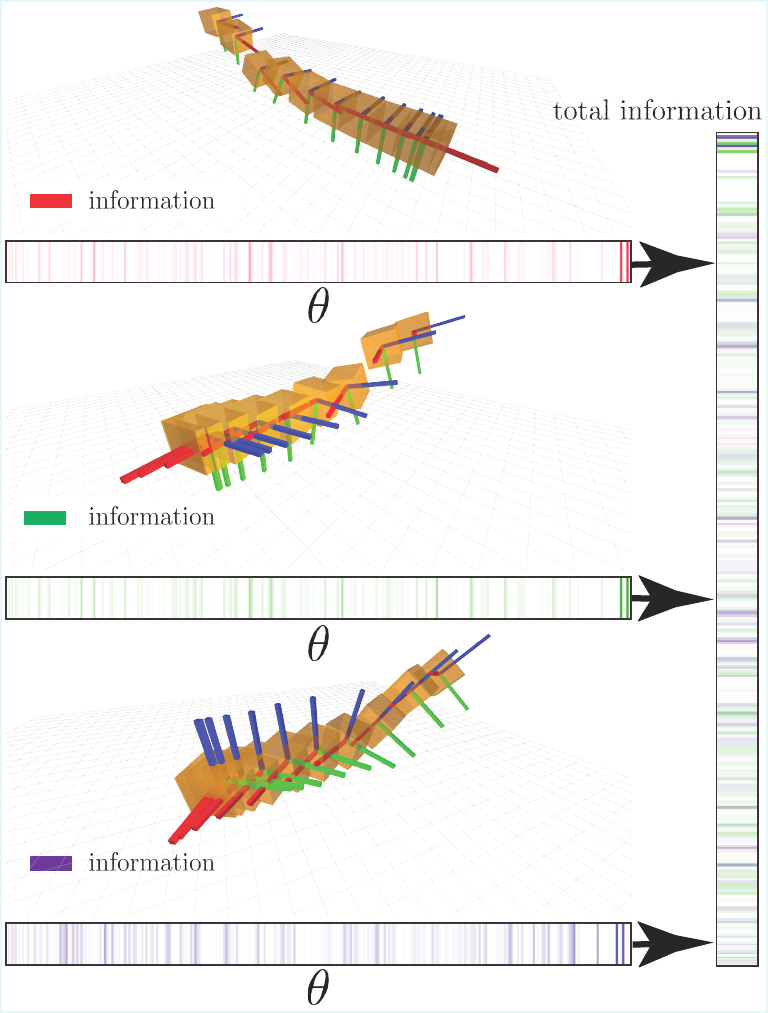}
        \caption{\textbf{Informative contact-data improves learning.} Here, we show demonstrations of block throws of varying measured trajectories $\tau$ (from real dataset \cite{ContactNets}) used to estimate unknown inter-body block signed distance functions. 
        % Different contact data reveals different information about parameters of interests. 
        Contact-based data collected from block trajectories contain varying amounts of information about parameters, indicated by color concentrations across the horizontal span of parameters of interest $\theta$. 
        Our goal is to regulate and rank contact data by its information content to quickly and efficiently converge on parameter uncertainty by populating maximal amount of total information (summation over all available information from the dataset) across all parameters.}
        \label{fig:abstract}
    \end{figure}
    
    We focus our study on curating data for improving learning from the ContactNets \cite{ContactNets} dataset. In~\cite{ContactNets}, the learning task seeks to obtain object physical and geometric properties directly through integrating contact impulse constraints.
    We show that a contact-aware formulation of the Fisher information \cite{sathyanarayan2025behaviorsynthesiscontactawarefisher} allows us to rank data used to train ContactNets based on its utility in improving the numerical conditioning of the learning problem.
    % posed as Linear Complementarity Problem (LCP).
    The data used for learning consists of tosses of a block that makes contact with a flat table. 
    We cast the objective in~\cite{ContactNets} as a maximum likelihood problem in order to derive measures of information that can be used to rank parameter-relevant data. 
    Empirically, our approach to data curation yields a data-efficient process for robot learning based on usefulness of data, rather than randomly sampling data for learning. 
    This paper presents a method that quantifies and ranks contact-aware Fisher information to regulate contact-based data used for learning by improving the conditioning of contact datasets to boost maximum likelihood parameter estimation problems \cite{ge2023maximumlikelihoodestimationneed}.
    For instance, in Fig. \ref{fig:abstract} we demonstrate independent block tosses that interact with a flat ground, where the objective is to estimate the block's physical parameters through contact interactions with the ground. 
    By searching for \emph{meaningful} contact interactions, we can avoid the presence of uninformative contact data for estimating parameters, thus regulating the information content in data with respect to parameters of interest.
    We also present a method that guides the learning problem even further by generating information-rich contact data to improve the parameter learning presented in \cite{ContactNets} using optimal experimental design methods \cite{fedorov2010optimal}.
    % We present a contact-aware optimal experimental design method that expands on the datasets in \cite{ContactNets} to improve the performance of parameter learning.

    Our contributions of this paper are the following:
    \begin{itemize}
        \item Formulation of a contact-aware Fisher information measure to curate and rank existing contact-based datasets presented in \cite{ContactNets}.
        % and \cite{chi2024diffusionpolicyvisuomotorpolicy}.
        \item Demonstrate the improved learning performance of our information-ranking method in comparison to conventional ContactNets.
        \item Pose a data generation method that can improve the initial conditions of block trajectory data generation to improve the contact-based learning performance of ContactNets.
    \end{itemize}

\section{RELATED WORK}

    \noindent\textbf{Parameter Estimation.}
    Commonly explored in \cite{MI_review,194738,10.5555/293154,nima}, parameter estimation involves fitting physical parameters of interests with a model structure that captures the dynamical behaviors of the robot, given a sequence of sensor measurements.
    While a variety of approaches exists, such as Bayesian inference \cite{PEPI2020103025} and factor-graph based methods \cite{sundaralingam2021inhandobjectdynamicsinferenceusing}, this paper formulates parameter estimation as a maximum likelihood problem \cite{ge2023maximumlikelihoodestimationneed}.
    The formulation allows us to explicitly reason over parameter estimation through sensor measurements, and is often used in many optimal experimental design techniques for optimal data generation.

    \noindent\textbf{Fisher Information and Experimental Design.} 
    Optimal experimental design \cite{AF_Emery_1998,fedorov2010optimal} offers methods to best sample data for estimating parameters via maximum likelihood.
    Generally, experimental design involves optimizing over an information measure to solve for \emph{expected} information-rich sensor data about unknown parameters of interest in an experiment .
    The Fisher information \cite{atanasov2013information,AF_Emery_1998} measures the sensitivity of measurements with respect to slight perturbations of parameters, and offers a lower bound on parameter uncertainty \cite{Rao_1947}.

    Fisher information naturally emerges as a byproduct to a Taylor expansion of the maximum likelihood problem \cite{ly2017tutorialfisherinformation}. 
    By maximizing over the Fisher information, we therefore condition the expected sensor measurements to contain information-rich data about parameters across a variety of robot parameter learning problems \cite{wilson_fishermax}.

    \noindent\textbf{Contact-Implicit Parameter Learning.}
    Continuous motion prediction models often fail to capture robotic systems undergoing contact with the environment.
    To tighten the model prediction error gap, sophisticated methods are often required to predict these non-smooth and discontinuous interactions.
    ContactNets \cite{ContactNets}, in particular, provides a smooth, contact-implicit learning method that learns inter-body signed distance functions and contact jacobians of dynamic systems given real trajectory data.
    Contact impulses are estimates through a relaxed, smoothened Linear Complementarity Problem (LCP) constraints that approximates contact interactions over non-convex sets.
    
    In~\cite{ContactNets}, it was found that physics-based structured learning technique outperform end-to-end methods when approximating parameters of interest.
    The train, test, and validation datasets used for training and testing their model are randomly shuffled and administered to their algorithm, and no further analysis of data quality was not considered in the paper. 
    In this paper, we directly consider information richness of the datasets given in ContactNets \cite{ContactNets} in order to rank and regulate the data provided to learning algorithm. 
    
    % \noindent\textbf{Diffusion Models.} 
    % Diffusion policies \cite{chi2024diffusionpolicyvisuomotorpolicy} present methods to learning a map state observations to actions as a conditional denoising diffusion model in robot action space.
    % By learning an action-score gradient conditioned over state observations from data collected from demonstrations, the authors present a model that has stable training, expresses arbitrarily multimodal action distributions, and scalable to high-dimensional output spaces.
    % In this work, we present a method that quantifies and ranks the information in data from demonstrations, which leads to significant improvements in learning.
    % The goal is to provide \emph{meaningful} data that maximizes the stochastic Langevin dynamics descent direction in action space to accelerate the policy learning process.
    
\section{PRELIMINARIES}

    \subsection{Parameter Estimation via Maximum Likelihood}
        Given a sequence of sensor measurements from the robot, our goal is to fit unknown parameters of interest to data by solving an optimization problem.
        In this paper, we utilize the following maximum likelihood estimation problem to solve for parameters of interest. 
        \begin{definition}\label{def:mle}
            \textbf{Maximum Likelihood Estimation.}
            Given parameters $\theta$, a dataset $\mathcal{D} = \{x_i\}_i^N$, where $x \in \mathcal{X}$ is an observation at instance $i$, for $N$ independent and identically distributed (i.i.d) measurements, and a likelihood measurement model $p\left(\mathcal{D} | \theta\right)$, the posterior distribution of the parameter estimate $\hat{\theta}$ is the solution to the optimization problem,
            \begin{equation}\label{eq:mle}
                \hat{\theta} = \arg\max_\theta \,\, \log p\left(\mathcal{D} | \theta \right) 
            \end{equation}
        \end{definition}

        % To massage numerical conditioning over the optimization, it is common practice to optimize over the log-likelihood model, $\arg\max_\theta \log p\left(\mathcal{D} | \theta \right)$.
        % \begin{remark}
        %     The assumption of gaussian measurements further simplifies the maximum likelihood problem to the following nonlinear least squares estimation problem,
        %     \begin{equation}
        %         \hat{\theta} = \arg\min_\theta \sum_{i=0}^N||x_i - f(x_i,\theta)||_{\Sigma^{-1}}
        %     \end{equation}
        %     where $f$ is the sensor measurment model that takes as inputs the parameters $\theta$ and observation $x_i$
        % \end{remark}

        The maximum likelihood estimation problem provides us with a convenient formulation that is useful for specifying experimental design problems.
        This paper utilizes the structure from maximum likelihood to examine the data utility existing datasets for learning, as well as generate the datasets $\mathcal{D}$ that best improves the conditioning of the optimization problem in \ref{eq:mle}. 
        
    \subsection{Fisher Information}

        To better condition the data for maximum likelihood estimation, we require a measure that directly considers the information-richness of the datasets with respect to parameters of interest. 
        Because certain datasets contain more information about parameters, we can select informative data to reduce parameter uncertainty.
        % Fisher information offers a Riemannian measure over statistical manifolds, and a direct correlation on information-richness of data when the gradient of the likelihood measurement model is large due to high parameter sensitivity.
        Fisher information offers a Riemannian measure over statistical manifolds, and the information content of data is proportional to larger score values of the log probabilities embedded in the Fisher information measure.
        \begin{definition}\label{def:fim}
        \textbf{Fisher Information Matrix.}
            Let $\mathcal{L}\left(\mathcal{D} | \theta \right) \in \mathcal{C}^2$ be the likelihood measurement model that is continuous and twice-differentiable over elements in dataset $\mathcal{D}$ and $\theta$. The Fisher Information Matrix  $\mathcal{F}\left(\mathcal{D}|\theta\right)$ models the outer product of the likelihood score function \cite{Rao1992},
            \begin{equation}
                \begin{split}
                    \mathcal{F}\left(\mathcal{D}|\theta\right)&=\mathbb{E}\left[\frac{\partial}{\partial \theta} \mathcal{L}(\mathcal{D} | \theta) \frac{\partial}{\partial \theta} \mathcal{L}(\mathcal{D} | \theta)^\top\right]\\
                    &=-\mathbb{E}\left[ \nabla_\theta^2 \mathcal{L}(\mathcal{D} | \theta) \right]
                \end{split}
            \end{equation}
        \end{definition}

        Under finite sampling over $N$ normally distributed observations comprised in dataset $\mathcal{D}=\{x_i\}_{i=1}^N$, we approximate the empirical Fisher information as,
        \begin{equation}\label{eq:empfim}
            \begin{split}
                \mathcal{F}(\mathcal{D} | \theta) &\approx - \frac{1}{N}\sum_{i=1}^N \frac{\partial^2 \mathcal{L}(x_i|\theta) }{\partial \theta^2} \\
                & = \frac{1}{N}\sum_{i=1}^N \frac{\partial \mathcal{L}(x_i | \theta)}{\partial \theta}
                \frac{\partial \mathcal{L}(x_i | \theta)}{\partial \theta}^\top.
            \end{split}
        \end{equation}
        
        % By selecting a dataset $\mathcal{D}$ that results in a maximally large likelihood score function signal, we have the ability to identify and regulate information-rich data to improve learning. 
        The Fisher information matrix also naturally arises as a byproduct from solving optimization problem \ref{eq:mle}.
        \begin{prop}
            Let $\mathcal{L}\left(\mathcal{D} | \theta \right)$ be the objective function of the maximum likelihood problem \ref{eq:mle}. Performing a second order Taylor expansion with respect to perturbations in parameter $\delta\theta$ around its unbiased estimator $\hat{\theta}$,
            \begin{equation}
                \delta\theta^* = \arg\max_{\delta\theta} \frac{\partial\mathcal{L}}{\partial\theta}\Big|_{\theta=\hat{\theta}}\delta\theta+\frac{1}{2}\delta\theta^\top \frac{\partial^2\mathcal{L}}{\partial\theta^2}\Big|_{\theta=\hat{\theta}}\delta\theta
            \end{equation}
            reveals a Hessian term that is directly related to the Fisher information matrix defined in Definition \ref{def:fim}.
        \end{prop}

        The Hessian term indicated above can be directly conditioned over using the Fisher information, which is a direct lower bound on parameter uncertainty.
        % By conditioning the data over the Fisher information matrix, we consequently improve the data quality used for parameter estimation that results in more precise parameter estimates, as Fisher information is a direct lower bound on parameter uncertainty.
        \begin{definition}\label{def:crlb}
            \textbf{Cram\'er-Rao Lower Bound \cite{Rao_1947}.}
            Given a likelihood distribution $p\left(\mathcal{D}|\theta\right)$, let $\mathcal{D}$ comprise randomly sampled observations from measurements. For an unbiased estimator $\hat{\theta}$, its variance is bounded by,
            \begin{equation}\label{eq:crlb}
                \mathcal{F}\left(\mathcal{D}|\theta\right) \succeq \textrm{cov}(\hat{\theta})^{-1}
            \end{equation}
        \end{definition}
        
        Therefore, maximizing optimality conditions over the Fisher information reveals diverse and meaningful \emph{expected} datasets that converge on parameter uncertainty.
        In this paper, we combine robot interaction-awareness with Fisher information as a measure to quantify information-richness in contact datasets to better improve contact-based parameter learning.
        
    \subsection{Optimal Experimental Design}\label{subsec:exp_design}

        We can improve the performance of maximum likelihood estimation by \emph{selecting} datasets $\mathcal{D}=\{x_i\}_{i=1}^N$ that maximize Fisher information, which is the crux of optimal experimental design. 
        We seek to improve the utility of the independent observation variable $x_i$ such that the resulting data collected from the experiment is conditioned to provide maximal information about unknown parameters of interest.
        \begin{definition}\label{def:oed}
        \textbf{Optimal Experimental Design.}
            Let $x_i\forall i\in[0,N]$ be an independent control variable denoting observations. The observations with maximal utility $x_i^*$ for estimating parameter $\theta \in \Theta$ is the solution to the optimal design,
            \begin{equation}
                \max_{x_i\forall i\in [0,N]} \Phi(\mathcal{F}(\mathcal{D}|\theta))
            \end{equation}
            where $\mathcal{D}=\{x_i,y(x_i)\}_{i=1}^N$ is a structured dataset of $N$ observations, and $\Phi(\cdot):\mathbb{R}^{\Theta \times \Theta} \rightarrow \mathbb{R}$ is a matrix reduction operator.
        \end{definition}

        In this paper, we utilize the \emph{contact-aware} Fisher information as both a ranking measure and a reward landscape to optimize for expected observations $x_i^*$. 
        The matrix reduction operator is used to project the information matrix to a scalar value, e.g determinant (used for D-optimality in experimental design), trace (T-optimality), or minimum eigenvalue (E-optimality).
        Because the Fisher information is symmetric and positive semi-definite \cite{ChirikjianGregoryS2009SMIT},
        we find that the reduction $\Phi(\cdot)$ has little variance in the outcome of the ranking quality.

\section{Curating Datasets via Contact-Aware Fisher Information}

    This section aims to derive the \emph{contact-aware} Fisher information as a measure to quantify and rank data utility for parameter learning problems.
    Our goal is to demonstrate the role of informative data in accelerating maximum likelihood estimation performance, as well as formulate a method that generates information-rich data relevant to parameter learning problems.

    \subsection{Contact-Aware Fisher Information as Rank Algorithm}

        The contact-aware Fisher information matrix captures how sensitive the parameters are to contact-based sensor data. 
        As aforementioned, such a measure can be directly derived by solving the maximum likelihood estimation problem.
        % \begin{definition}\label{def:cmap}
        %     \textbf{Contact-Aware Maximum Likelihood Problem.}
        %     Let $\mathcal{D}=\{\tau\}_{j=1}^K$, where $\tau=\{(x_i,\lambda_i)\forall i \in [0,T]\}$ is a trajectory sequence of robot states $x_i\in\mathcal{X}$ contact impulses $\lambda\in\mathcal{C}$ over some finite trajectory length $T\in \mathbb{Z}^+$, and $\theta$ be the unbiased parameter estimate of $\hat{\theta}$. Given a gaussian likelihood measurement model $p(\mathcal{D}|,\theta)$, where $g_\theta = \mathbb{E}_{\tau \sim \mathcal{T}}\Big[p(\mathcal{D}|\theta)\Big]$, the contact-aware maximum likelihood problem is defined as,
        %     \begin{equation}\label{eq:cmap}
        %         \begin{aligned}
        %             & \  
        %             \hat{\theta} = \arg\max_{\theta, \tau} \log p(\mathcal{D} | \theta)  \\ 
        %             \text{s.t.} & 
        %             \begin{cases}
        %                 x_0 \text{ given } & \text{ (initial state) } \\ 
        %                 x_{t+1} = f_\theta(x_t, u_t, \lambda_t) & \text{ (dynamics) } \\ 
        %                 \lambda_t \in \mathcal{C}_\theta(x_t) & \text{ (contact constraints) }
        %             \end{cases} \\
        %         \end{aligned}
        %     \end{equation}
        % \end{definition}
        The Fisher information matrix is a natural byproduct of solving for optimality conditions over the maximum-likelihood estimation problem in eq. \ref{eq:mle}.
        % \begin{equation}\label{eq:cfim}
        %     \begin{split}
        %         \mathcal{F}(\mathcal{D}|\theta)&\approx-\nabla_\theta^2\mathcal{L}(\mathcal{D}|\theta) \\
        %         &= \nabla_\theta\mathcal{L}(\mathcal{D}|\theta)\otimes\nabla_\theta\mathcal{L}(\mathcal{D}|\theta)^\top
        %     \end{split}
        % \end{equation}
        % where $\mathcal{L}$ is the objective function of the maximum likelihood problem. 
        % For the full proof of how contact-aware Fisher information is revealed from the optimization problem \ref{eq:cmap}, see the proof in.

        Specifically to ContactNets, we compute the Fisher information as an outerproduct of the score of the loss function detailed in \cite{ContactNets}. 
        The authors smoothen the discontinuities and non-smoothness of contact by including implicit smoothing over the contact impulses, resulting in more well-behaved contacts.
        Taking the gradient over the objective yields the following ContactNets-specific score function,
        \begin{equation}
            \begin{split}
                \nabla_\theta \mathcal{L}(\mathcal{D}|\theta) &= 
                2 \sum_{i=1}^N \left(\nabla_\theta {\mathcal{J}_i^\theta}^\top \lambda_i \right)^\top \left({\mathcal{J}_i^\theta}^\top \lambda_i - f_{data,i}\right)
                \\
                &+ 2\nabla_\theta {\phi_{n,i}'^\theta}^\top {\phi_{n,i}'^\theta} ||\lambda_i||^2 
                + \nabla_\theta\min (0, {\phi_{n,i}'^\theta})^2 \\
                &+ 2 \left( \frac{\nabla_\theta \mathcal{J}_{t,i}^\theta}{||\mathcal{J}_{t,i}^\theta v'||_2} v' \lambda_{t,i} + \lambda_{n,i} \nabla_\theta \mathcal{J}_{t,i}^\theta v'\right)^\top \\ 
                & \ \ \ \ \ \ \ \ \ \ \ \ \ \ \ \ \left( ||\mathcal{J}_{t,i}^\theta v'||_2 \lambda_{t,i} + \lambda_{n,i} \mathcal{J}_{t,i}^\theta v' \right)
            \end{split}
        \end{equation}
        where $f_{data}$ is the contact-based dataset used for learning, $\mathcal{J}_i^\theta = \left[\mathcal{J}_{n,i}^\theta, \mathcal{J}_{t,i}^\theta\right]$ are the normal and tangent contact Jacobians, $\lambda_i$ is the contact impulse, $\phi_{n,i}'^\theta$ and $v'$ are  next-step predictions of the signed distance function between the block and ground and block velocity, respectively, governed by an explicit Euler step trajectory prediction as detailed in \cite{ContactNets}.
        
        This paper utilizes the contact-aware Fisher information measure to quantify the data utility of real contact-based data and ranks the data quality using a matrix reduction $\Phi(\cdot)$ over the Fisher information.
        % \begin{figure}
        %     \centering
        %     \includegraphics[width=0.9\linewidth]{figs/results/diagram.jpg}
        %     \caption{Caption}
        %     \label{fig:placeholder}
        % \end{figure}
        \begin{algorithm}
            \caption{Contact-Aware Fisher Dataset Ranker}\label{alg:rank}
            \begin{algorithmic}
                \Require dataset $\mathcal{D}$, matrix reduction operator $\Phi(\cdot)$, number of datasets $k_{thresh}$, number of total datasets $k_{total}$.
                \State \textbf{Initialize:} $k=0$, empty list $\Gamma$
                \While{$k< k_{total}$}
                    \State $\mathcal{F} \gets \textrm{Contact-Aware Fisher Inf. (eq. \ref{eq:empfim})}$

                    \State Append $\Phi(\mathcal{F})$ to $\Gamma$\Comment{append/compute info.}
                    \State $k \gets k+1$
                \EndWhile
                \State $\mathcal{S} \gets \textrm{argsorted}(\Gamma)$\Comment{sort arguments} \\
                \Return $\textrm{flip}(\mathcal{S}[-k_{thresh}:])$\Comment{return select indices}
            \end{algorithmic}
        \end{algorithm}
        Algorithm \ref{alg:rank} overviews the ranking algorithm implemented with the contact-aware Fisher information measure.
        We seek to sample observations from the set $\mathcal{D}$ that hold maximal utility to the maximum likelihood estimation problem.
        This algorithm returns ranked indices that correspond to the order of the dataset stack up to certain number of datasets of choice $k_{thresh}$.
        
        % To contextualize the notion of Fisher-ordered ranking, consider $\mathcal{D}$ to be a dataset sampled from the measurement likelihood model $p(\mathcal{D}|\tau,\theta)$. The set $\mathcal{D}$ is a ranked dataset by data utility with respect to parameters of interest $\theta$ if the following inequality holds,
        %     \begin{equation}\label{eq:fishrank}
        %         \mathcal{D}_k \leq_{\Phi(\mathcal{F})} \mathcal{D}_{k+1} \ \forall k \in [0,K-1]
        %     \end{equation}
        % where $\leq_{\Phi(\mathcal{F})}$ denotes an inequality projection over a scalar-reduced information space from the mapping $\Phi(\cdot):\mathbb{R}^{\Theta\times\Theta} \rightarrow \mathbb{R}$.
        % \begin{definition}\label{def:fishrank}
        %     \textbf{Fisher-order Data Ranking.}
        %     Let $\mathbf{D}=\{\mathcal{D}_1,\mathcal{D}_2,\cdots,\mathcal{D}_K\}$ be a set of datasets of observations sampled from the measurement likelihood model $p(\mathcal{D}|\tau,\theta)$. The set $\mathbf{D}$ is a ranked dataset by data utility with respect to parameters of interest $\theta$ if the following inequality holds,
        %     \begin{equation}\label{eq:fishrank}
        %         \mathcal{D}_k \leq_{\Phi(\mathcal{F})} \mathcal{D}_{k+1} \ \forall k \in [0,K-1]
        %     \end{equation}
        %     where $\leq_{\Phi(\mathcal{F})}$ denotes an inequality projection over a scalar-reduced information space from the mapping $\Phi(\cdot):\mathbb{R}^{\Theta\times\Theta} \rightarrow \mathbb{R}$.
        % \end{definition}
        By selecting the top $k_{thresh}$ number of information-rich contact datasets for learning parameters, we converge on parameter uncertainty quicker than random selection of data.
        
\begin{figure*}[t!]
    \centering
    \makebox[\textwidth][c]{%
    \hspace{1cm}
    \includegraphics[width=1.2\linewidth]{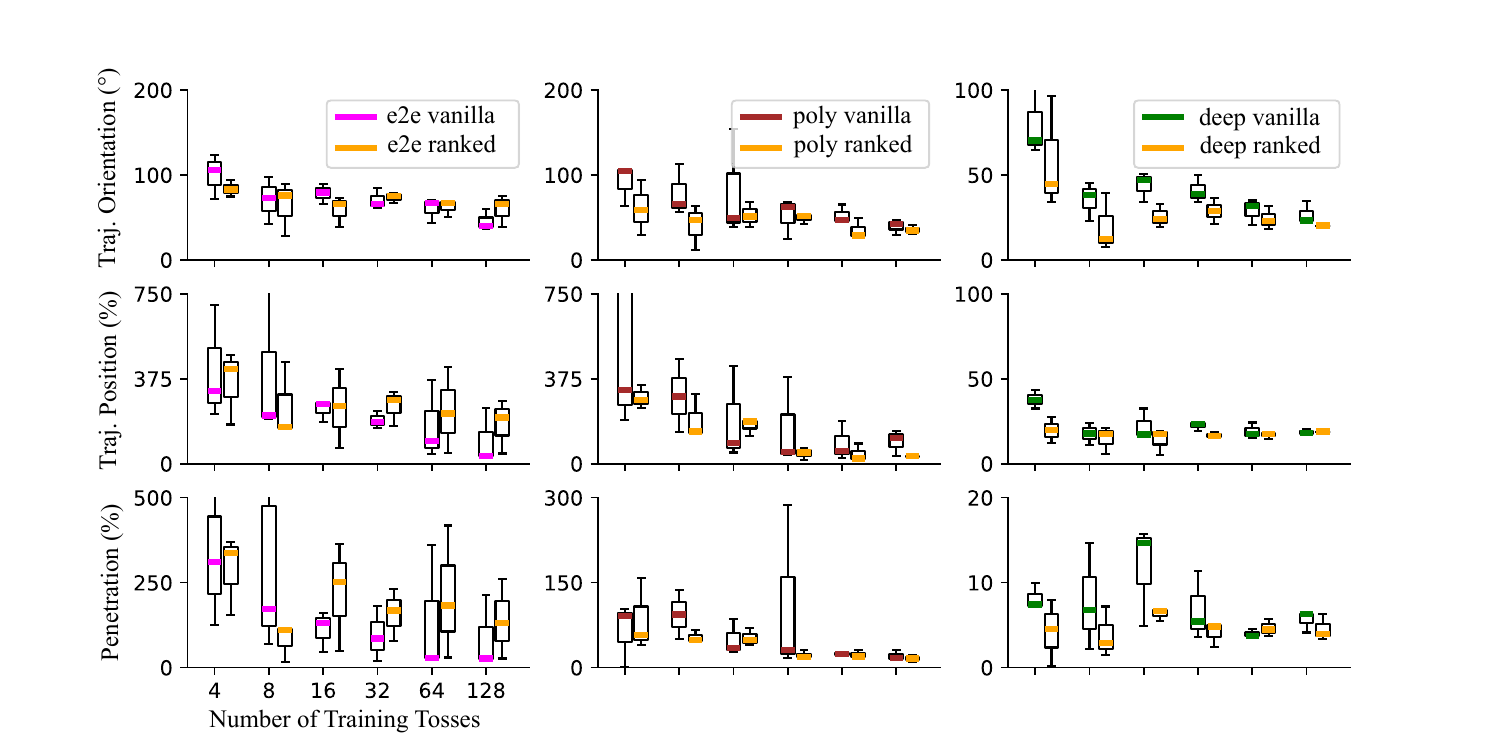}%
  }
    \caption{\textbf{Dataset Quality Improves Performance. } Here, we show the performance over three models written in \cite{ContactNets} for three selected performance metrics over fixed numbers of data sizes. We observe overall improvement of our traced Fisher-ordered ranking approach across all methods evaluated across three samples per experiment of fixed data size. We observe weaker improvement over end-to-end experiments due to the degeneracy of gradients of fitting a network directly to the contact impulses. By contrast, we see stronger improvement, on average, for the ContactNets polytope and deepvertex ranked methods due to the explicit physics-based structure applied to both models, significantly improving parameter sensitivity gradients. Additionally, we show, on average, that our approach generates lower variance of trajectory and penetration errors over the sampled experiments.}
    \label{fig:losses}
\end{figure*}

    \subsection{The Impact of (Un)-Informative Data}

        This section aims to show the importance of informative contact data for robot parameter learning problems, and, conversely, the drawbacks of random selection of data.
        Through direct reasoning over the nature of contact, we can directly correlate robot motion and interactions with the environment to its resulting informativeness about unknown parameters of interest via the contact-aware Fisher information measure.

        \begin{theorem}
            Let $\mathcal{D}$ be a dataset of observations sampled from the likelihood model $p(\mathcal{D}|\tau,\theta)$. Then,
            \begin{align*}
                \exists N \leq |\mathcal{D}| \in \mathbb{R} \ \ \ \textrm{s.t.} \ \ \mathcal{I} \approx \mathcal{\mathcal{F}(\mathcal{\tilde{D}}|\theta)}^{-1} \mathcal{F}(\mathcal{D}|\theta)
            \end{align*}
            where,
            \begin{align*}
                |\tilde{\mathcal{D}}| = N
            \end{align*}
        \end{theorem}
        \begin{proof}
            Let $\theta$ be an unbiased estimator of $\hat{\theta}$, $x_i \sim \mathcal{D}$ be a sampled datapoint, and let its corresponding deterministic measurement score at the sample be,
            \begin{align*}
                g_i = \nabla_\theta \log(p(x_i|\theta))
            \end{align*}
            and the Fisher information matrix for the full dataset is,
            \begin{align*}
                \mathcal{F}=\sum_{i=1}^{|D|} g_i g_i^\top
            \end{align*}
            If there exists a subset $\mathcal{\tilde{D}}\subset\mathcal{D}$ such that
            \begin{align*}
                \lim_{n\rightarrow N} \sum_{i=1}^{|D|} g_i g_i^\top - \sum_{i=1}^{N} g_i g_i^\top \approx 0
            \end{align*}
            then the corresponding Fisher information $\mathcal{F}(\mathcal{\tilde{D}}|\theta)$ satisfies the bounds,
            \begin{align*}
                (1-\epsilon)\mathcal{F} \leq \mathcal{\tilde{F}} \leq (1+\epsilon)\mathcal{F}
            \end{align*}
            for arbitrarily small $\epsilon$.
            Because $\mathcal{F}, \mathcal{\tilde{F}}$ are positive, semidefinite matrices \cite{pmlr-v70-sun17b} and assumed diagonal due to isotropic sensor noise, the corresponding eigenvalues $\Lambda \in \textrm{eig}(\mathcal{\tilde{F}}^{-1}\mathcal{F})$ satisfy the bounds,
            \begin{align*}
                \frac{1}{1-\epsilon} \leq \lambda \leq \frac{1}{1+\epsilon} \ \ \ \forall \lambda \in \Lambda
            \end{align*}
            Therefore, the following inequality is satisfied,
            \begin{align*}
                ||\mathcal{\tilde{F}}^{-1}\mathcal{F} - \mathcal{I}|| \leq \frac{\epsilon}{1+\epsilon}
            \end{align*}
            As $\epsilon \rightarrow 0$, and $\mathcal{\tilde{F}}^{-1}\mathcal{F} \succeq 0$,
            \begin{align*}
                \mathcal{\tilde{F}}^{-1}\mathcal{F} \approx \mathcal{I} 
            \end{align*}
            thus completing the proof. 
            % \qed
        \end{proof}
        We essentially show above that a smaller subset of data used for learning is sufficient if the information content in the data is Fisher-informative with respect to parameters.
        
        The role of the matrix reduction operator $\Phi(\cdot)$ plays defines the informativeness of a dataset.
        However, the choice of matrix reduction operators is rather insignificant under certain regularity conditions \cite{10.1214/aos/1176342810}, especially under high data regimes.
        The rationale for this is largely true if the dataset used for learning is diverse and informative to a wide spectrum of parameters of interests.
        However, under lower quantities of data, simply ranking based on, for instance, the trace of the Fisher information can lead to poor performance of downstream learning if the top-ranked datapoints are all maximally information-rich for the same or similar subspace of parameters.

        To circumvent this challenge, we present a novel ranking method that ranks the informativeness of data by observing the \emph{orthogonality} of each of the dataset's likelihood measurement model score function with respect to one another.
        
        For instance, let $\mathcal{D}=\{\tau\}_{i=1}^N$ be a dataset trajectory measurements from individual experiments.  
        Two datapoints with maximally orthogonal information with respect to parameters of interest $\theta$ can be found as a solution to,
        \begin{equation}
            \min_{\tau_{i,j}} \ | \nabla_\theta \log \ p\left(\tau_j|\theta\right) \cdot \nabla_\theta \log \ p\left(\tau_k|\theta\right)| \ \ \textrm{where } \tau_j , \tau_k \sim \mathcal{D}
        \end{equation}
        The core idea is to measure the \emph{dissimilarity} of contact information from one dataset to another. 
        By measuring `info-orthogonality', we can better rank datasets that sparsely communicate information across different subspaces of parameters.

        % \begin{remark}
        %     If set $\mathcal{D}$ is a ranked dataset of data utility with respect to parameters of interest $\theta$, then the following inequality is true,
        %     \begin{equation}
        %         \mathcal{F}(\mathcal{D}_{k+1}|\theta) - \mathcal{F}(\mathcal{D}_k|\theta) \succeq 0 \ \ \forall k \in [0,K-1]
        %     \end{equation}
        %     and consequently, for any linear operator $\Phi(\cdot)$, 
        %     \begin{equation}
        %         \Phi\left(\mathcal{F}(\mathcal{D}_{k+1}|\theta)\right) - \Phi\left(\mathcal{F}(\mathcal{D}_k|\theta)\right) \geq 0
        %     \end{equation}
        % \end{remark}

    \subsection{Data-Generation via Contact-Aware Fisher Information}

        We show how the contact-aware fisher information can be used to improve experiments.
        We expand on existing datasets presented in ContactNets \cite{ContactNets} by optimizing new contact datasets governed by contact-aware experimental design techniques.
        \begin{definition}\label{def:caoed}
            \textbf{Contact-Aware Optimal Experimental Design.}
            Let $\tau=\{(x_i,\lambda_i)\forall i\in [0,T]\}$, where $x \in \mathcal{X}$ is the robot state and $\lambda \in \Lambda$, be a sequence of $T$ state-contact impulse observations, $\Phi(\cdot):\mathbb{R}^{\Theta\times\Theta}\rightarrow\mathbb{R}$ be a continuous and differentiable map over a positive, semi-definite Fisher information matrix $\mathcal{F}(\mathcal{D}|\theta)\in\mathbb{R}^{\Theta\times\Theta}$ \cite{pmlr-v70-sun17b}. For a given parameter estimate $\theta\in\Theta$, the general contact-aware experimental design problem is defined as,
            \begin{align} \label{eq:caoed}
                    \hat{\mathcal{D}} &= \argmax_{\mathcal{D}} \Phi\left(\mathcal{F}(\mathcal{D}| \theta)\right)\\ &\text{ s.t. }
                    \begin{cases}
                        x_0 \sim p(x_0) & \text{(sampled initial state)}\\
                        x_{t+1} = f_\theta(x_t,u_t,\lambda_t) & \textrm{(dynamics constraint)} \\
                        \lambda_t \in \Lambda & \textrm{(contact constraint)}
                    \end{cases} \nonumber
                \end{align}
                where $p(x_0)$ is a normally distributed model over the initial conditions.
        \end{definition}
        By optimizing over the Fisher information measure, we condition diverse and meaningful datasets that are utilized to maximally improve upon parameter precision during the parameter learning process.
        \begin{algorithm}
        \caption{Contact-Aware Data Generator}\label{alg:caoed}
            \begin{algorithmic}
                \Require initial conditions $x_0$, initial state distribution $p(x_0)$, parametrized dynamics, sensor, and contact models, trajectory length $T$, numer of experiments $N_{exp}$
                \State \textbf{Initialize: }$n=0$, empty list $\mathcal{D}_{total}$
                \While{$n < N_{exp}$}
                    \State $\hat{\mathcal{D}}\gets$ CA-OED Eq.~\eqref{eq:caoed} \Comment{solve exp. design}
                    \State $\mathcal{D}\gets$ Execute experiment $\hat{\mathcal{D}}$ on robot
                    \State Append $\mathcal{D}$ to $\mathcal{D}_{total}$
                    \State $n \gets n + 1$
                \EndWhile
            \end{algorithmic}
        \end{algorithm}

\begin{figure*}[t!]
    \centering
    \includegraphics[width=\textwidth]{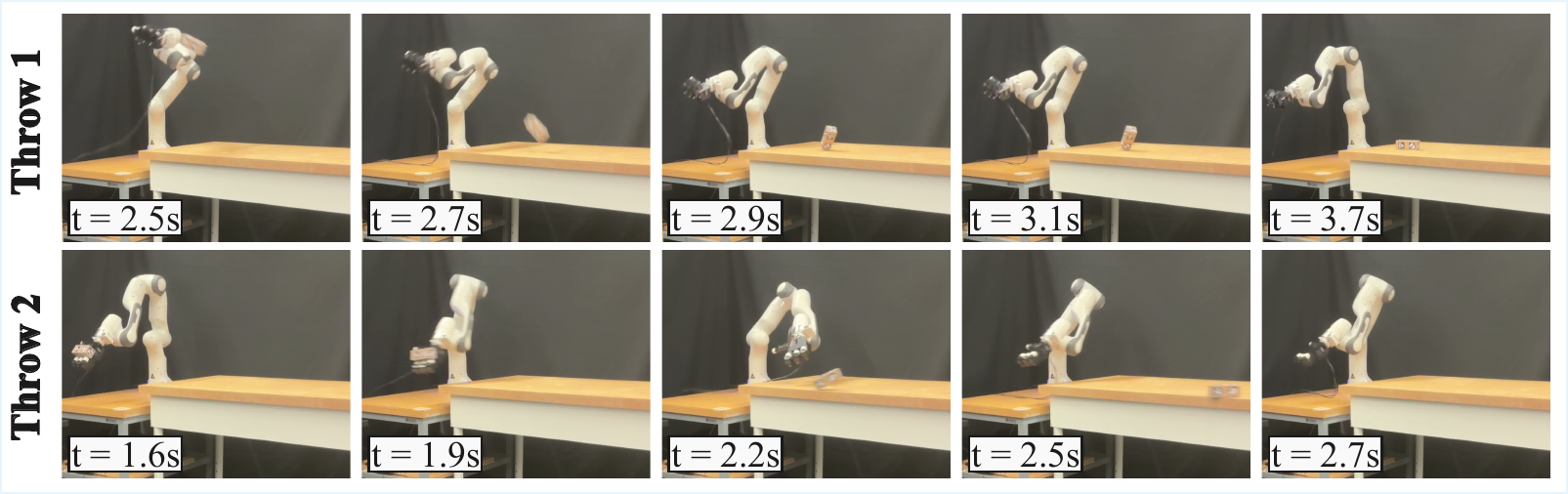}
    \caption{\textbf{Experimental Design for Learning. } We show two different robot demonstrations from our proposed contact-aware experimental design approach. The robot is tasked with selecting optimal initial throws that result in information-rich contact data (block trajectory measured by AprilTags). In the first throw, we indicate that improving on the initial conditions of optimal throws results in high excitations in rotation from contact, thus flipping the block over upon contact to identify inter-body signed distance functions for all vertices of the block. In the second throw, we indicate that our method excites information-rich contacts through high sliding forces on the block for proper identification of tangent contact Jacobians.}
    \label{fig:traj_examples}
\end{figure*}

\section{RESULTS}

    We demonstrate the effectiveness of our Fisher-based data ranking algorithm by implementing to ContactNets \cite{ContactNets}.
    % and a Push-T diffusion model \cite{chi2024diffusionpolicyvisuomotorpolicy}.
    Additionally, we test our data generation approach on ContactNets by generating contact data on a block throwing system via active contact exploration.

    ContactNets seeks to learn inter-body signed distance functions and contact Jacobians of a thrown block interacting with a flat ground.
    Contact data in this case is a trajectory sequence of states measured from AprilTag sensors attached to the block.
    Using the available data provided in the ContactNets repository, we evaluate ContactNets over at least 12 epochs of training for three methods:

    \noindent\textbf{Polytope.}
    A structured geometric model is provided to the learnable signed distance function $\phi_{poly}(q)$, where the block is approximated as a set of vertices that are the only entities that make contact with the ground.
    
    \noindent\textbf{DeepVertex.}
    The polytope model is augmented with an additional Deep Neural Network (DNN) term to account for arbitrary object and ground geometries,
    \begin{equation}
        \phi_{deep}(q)=\phi_{poly}(q)+\phi_{DNN}^\theta(q)
    \end{equation}

    \noindent\textbf{End-to-end.}
    A DNN is fitted directly on the observed contact forces to relieve the unstructured model from solving the contact impulses,
    \begin{equation}
        \mathcal{L}_{e2e}=||f_{data} - f_{DNN}^\theta(x,u)||_2^2
    \end{equation}

    All learnable parameters are outlined in Appendix A.2 in \cite{ContactNets}.
    Our goal is to augment the existing algorithms for these approaches by incorporating our ranking algorithm \ref{alg:rank} for every $n$ epochs of training over fixed data sizes per experiment.
    The core idea is that through careful curating of the contact data through our Fisher-ranking measure, we converge on parameter uncertainty faster than uniform random sampling of datasets for training, validation, and testing.
    We evaluate the performance of our approach by measuring the error of our learned parameters with test data.

    % We also evaluate the task completion of a push-T task after training a state-based diffusion policy \cite{chi2024diffusionpolicyvisuomotorpolicy} for {\color{red} n} epochs.
    % Our Fisher information is modeled by computing the action gradient outer product after every {\color{red} m} epochs, which is then used as a measure to rank the existing datasets.

\begin{figure}
    \centering
    % \vspace{-0.7cm}
    \includegraphics[width=\linewidth]{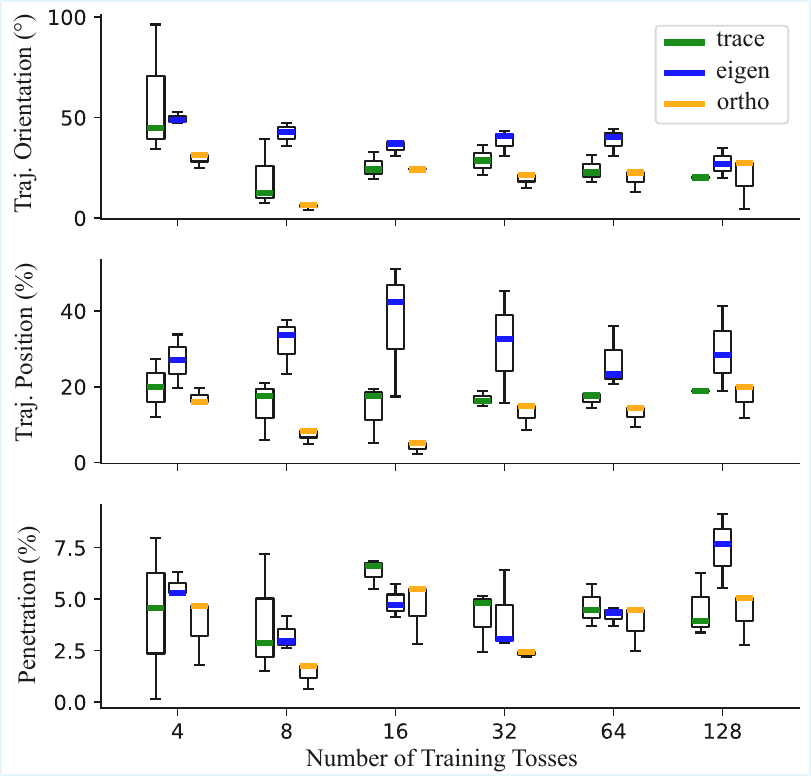}
    \caption{\textbf{Fisher Functionals Invariably Improve Learning} Here, we show results of our Fisher-order ranking method using different functional $\Phi(\cdot)$ mappings: trace, minimum eigenvalue, and our info-orthogonal ranking method. All methods reveal slightly different, but similar, extractions of Fisher information for the curation of data for learning. However, we anticipate that under the presence of datasets with largely diverse informativeness of parameters, the info-orthogonal methods improve the data ranking quality under low-data regimes.}
    \label{fig:functionals}
\end{figure}

    \subsection{Dataset Quality Improves Performance}

        We show the performance of our Fisher-ordered data ranking algorithm in comparison to methods from ContactNets \cite{ContactNets} in Fig. \ref{fig:losses}. 
        We evaluate the performance of all methods using performance metrics described in \cite{ContactNets}. 
        
        Significant improvements are observed in error reduction across all three Fisher-order ranking methods, with stronger improvements for the ranked ContactNets Polytope and Deepvertex methods. 
        Weaker improvements for end-to-end methods are attributed to the degeneracy of available gradients through the contact, as end-to-end methods \emph{directly} map a neural network to contact impulses. 
        In contrast, DeepVertex has access to a rich gradients from both the signed distance function written for Polytope as well as an additional overparametrized neural network to resolve complex contact interactions, thus converging on model uncertainty quicker.

\begin{figure}
    \centering
    % \vspace{-0.7cm}
    \includegraphics[width=\linewidth]{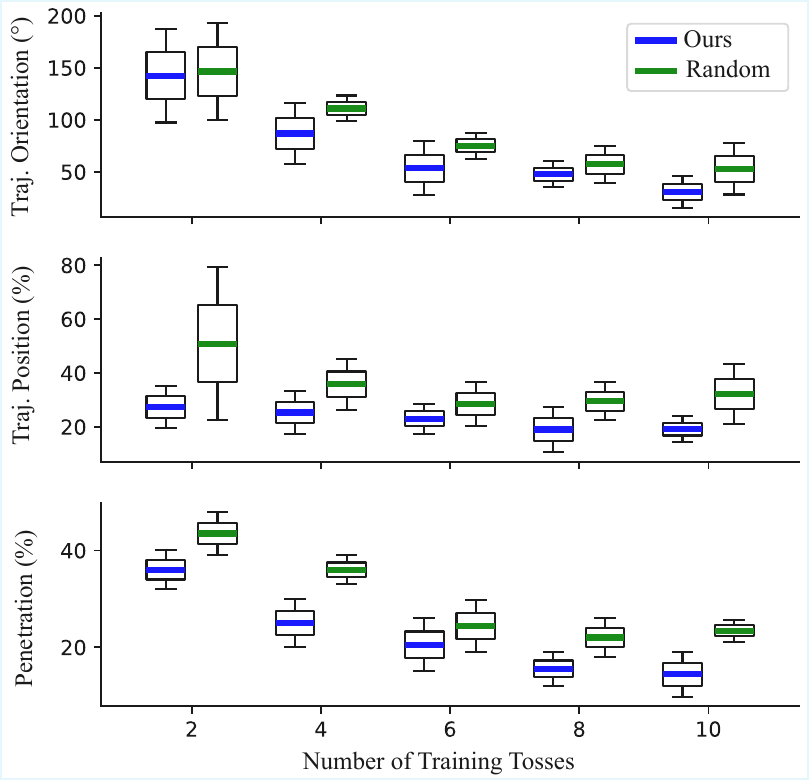}
    \caption{\textbf{Optimal Data Generation via Fisher Information Maximization. } 
    % Here, we demonstrate our Fisher information maximization method by generating optimal contact data across fixed number of data sizes used for training.
    Using our algorithm, we optimize over the contact-aware Fisher information matrix to generate an initial block throw that generates \emph{expected} information-rich contact data used for learning.
    We utilize the trace of the Fisher information as our matrix reduction operator.
    Our approach is compared with a normal random sampling method that samples random initial block throws and executes the throw on the real robot.
    The mean and variance of the normal distribution used to sample initial throws is calculated empirically from the existing ContactNets dataset.
    % We generate the contact data using our approach prior to training over the ContactNets DeepVertex models.
    Our results show overall improvement by explicitly reasoning over the information in contact compared to random sampling, thus boosting the ContactNets DeepVertex learning process.
    }
    \label{fig:hardware_results}
\end{figure}

    \subsection{Fisher Functional Invariance to Ranking}

        The results of our info-orthogonal approach are shown in Fig. \ref{fig:functionals}.
        We observe that under low data regimes, our info-orthogonal method outperforms conventional functionals over the Fisher information, trace and minimum eigenvalue.
        However, as the data size increases, the performance across all the methods are insignificant to each other. 
        This behavior is expected if the dataset used for training is sufficiently diverse (in the sense of Fisher information) across a wide spectrum of parameters of interest.

    \subsection{Accelerating Parameter Learning via Contact Dataset Generation}

    We present results on conditioning the contact datasets used for learning by actively exploring for information-rich data prior to the training process.
    By generating information-rich data for learning unknown parameters of interest, we consequently converge on parameter uncertainty quicker compared to na\"ive random sampling of initial conditions for data gathering.
    Particular to ContactNets, we aim to find optimal initial throws that maximizes the contact-aware Fisher information with respect to the parameters.

    The resulting trajectory errors are shown in Fig. \ref{fig:hardware_results}, with example trajectories from our experimental design method shown in Fig. 
    \ref{fig:traj_examples}.
    We compare our approach to a random sampling-based method, where an initial throw is sampled from a normal gaussian distribution, with mean and variances calculated empirically from the existing ContactNets data.
    Our experimental design method also samples from this distribution, however it computes and optimizes over a Fisher information cost function that chooses samples that are information-rich with respect to unknown parameters of interest to estimate inter-body signed distance functions and block-ground contact Jacobians.
    We show that our method outperforms random sampling, and selecting the initial conditions for generating contact data \emph{directly} yields proper conditioning of datasets to converge on parameter uncertainty quicker.

\section{CONCLUSIONS}

    We demonstrate a method that curates and ranks contact datasets to improve the learning performance of ContactNets.
    We show that careful regulation over datasets directly yields information-rich data for learning unknown parameters of interest.
    We also extend our approach by formulating an experimental design method that optimizes over the initial conditions to generate expected information-rich contact data to improve the learning performance compared to random sampling methods.

\addtolength{\textheight}{-10cm}   % This command serves to balance the column lengths

%%%%%%%%%%%%%%%%%%%%%%%%%%%%%%%%%%%%%%%%%%%%%%%%%%%%%%%%%%%%%%%%%%%%%%%%%%%%%%%%

%%%%%%%%%%%%%%%%%%%%%%%%%%%%%%%%%%%%%%%%%%%%%%%%%%%%%%%%%%%%%%%%%%%%%%%%%%%%%%%%



% \section*{ACKNOWLEDGMENT}

% The preferred spelling of the word ÒacknowledgmentÓ in America is without an ÒeÓ after the ÒgÓ. Avoid the stilted expression, ÒOne of us (R. B. G.) thanks . . .Ó  Instead, try ÒR. B. G. thanksÓ. Put sponsor acknowledgments in the unnumbered footnote on the first page.
\bibliographystyle{IEEEtran}
\bibliography{root}

\end{document}